\documentclass[sn-mathphys, iicol]{sn-jnl}

\DeclareMathOperator{\EX}{\mathbb{E}}

\jyear{2021}%

\theoremstyle{thmstyleone}%
\theoremstyle{thmstyletwo}%

\theoremstyle{thmstylethree}%

\usepackage{graphicx}
\usepackage{caption}
\usepackage{subcaption}
\usepackage{longtable} 
\usepackage{enumitem} 
\usepackage{array}
\usepackage{gensymb} 
\usepackage{textcomp} 
\usepackage{pifont} 
\usepackage{numprint}
\usepackage{amsmath,amssymb,amsfonts}
\usepackage{adjustbox}
\usepackage{fixme}
\usepackage{hyperref}
\fxsetup{status=draft}

\begin{document}

\title[Latent space of generative models]{Comparing the latent space of generative models}


\author*[1]{\fnm{Andrea} \sur{Asperti}}\email{andrea.asperti@unibo.it}

\author*[1]{\fnm{Valerio} \sur{Tonelli}}\email{valerio.tonelli2@studio.unibo.it}


\affil*[1]{\orgdiv{Department of Informatics: Science and Engineering (DISI)}, \orgname{University of Bologna}, \orgaddress{\street{Mura Anteo Zamboni 7}, \city{Bologna}, \postcode{40126},
\country{Italy}
}}





\abstract{Different encodings of datapoints in the latent space of latent-vector generative models may result in more or less effective and disentangled characterizations of the different explanatory factors of variation behind the data. Many works have been recently devoted to the exploration
of the latent space of specific models, mostly focused on the study of how features are disentangled and of how trajectories producing desired alterations of data in the visible space can be found. In this work we
address the more general problem of comparing the latent spaces of different models,
looking for transformations between them. We confined the investigation to the familiar
and largely investigated case of generative models for the data manifold of human faces.
The surprising, preliminary result reported 
in this article
is that (provided models have not been taught or explicitly conceived to act differently)
a simple linear mapping is enough to pass from a latent space to another while
preserving most of the information.}


%
%
%

\keywords{Generative Models, Latent Space, Representation Learning, Generative Adversarial Networks, Variational Autoencoders}


\maketitle

\section{Introduction}\label{sec1}
The task of generating new data from samples has always exerted a particular fascination in machine learning, both because of the potential for almost endless streams of new and original data, as well as for the implications on the knowledge extracted by 
a model about the data manifold. It is clear that the effectiveness of generative techniques
crucially depends on data representation, and different encodings may result in 
more or less entangled combinations of the different explanatory factors of variation 
behind the data \cite{BengioCV13,KimMnih18}. The key idea behind unsupervised 
learning of disentangled representations is that real-world data depends on a relatively
small number of explanatory factors of variation which can be compressed and recovered by unsupervised learning techniques \cite{Locatello19,helpful19,Locatello20}. Strictly related to 
representation learning, the task of {\em exploration} of the latent space of 
generative models aims to understand the ``arithmetic" of the variational factors \cite{Radford15,Closed-Form-fact}, and 
the effect that particular trajectories inside the latent space could produce in the visible
domain \cite{Shen22, learning_latent18, li2021discovering}.

\begin{figure}[h]
\begin{center}
\includegraphics[width=\columnwidth]{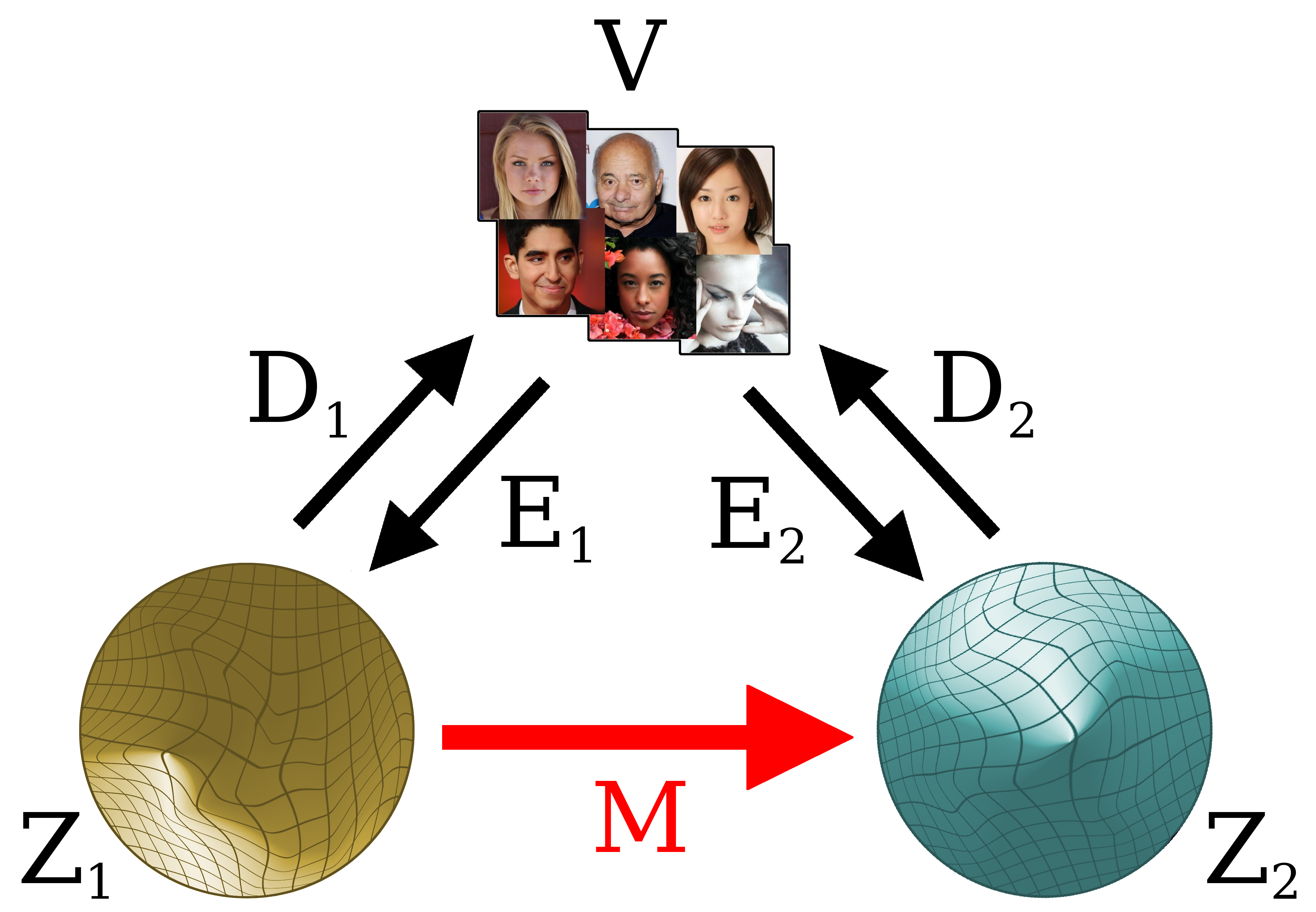}
\caption{
\label{fig:mapping} Given a generative model, it is usually possible to have an encoder-decoder pair mapping the visible space to the latent one (even GANs can be inverted, see Section \ref{sec:inversion}). 
From this assumption, it is always possible to map an internal representation in a space $Z_1$ to the corresponding internal representation
in a different space $Z_2$ by passing through the visible domain. This provides a supervised set of input/output pairs: we can try to learn a direct map, as simple as possible.
The astonishing fact is that a simple linear map gives excellent results, in many situations. This is quite surprising, given that 
both encoder and decoder functions are modeled by deep, non-linear
transformations.}
\end{center}
\end{figure}

In spite of the huge amount of work devoted to the exploration of latent spaces, relatively
little attention has been so far devoted to the problem of {\em comparing} the latent space
of different generative techniques, i.e. to the problem of locating the internal 
representation $z_X$ of $X$ in a given space starting from its representation in the latent space of a different model (see Figure \ref{fig:mapping}).

\begin{figure*}
     \centering
     \begin{subfigure}[b]{\textwidth}
         \centering
         \includegraphics[width=\textwidth]{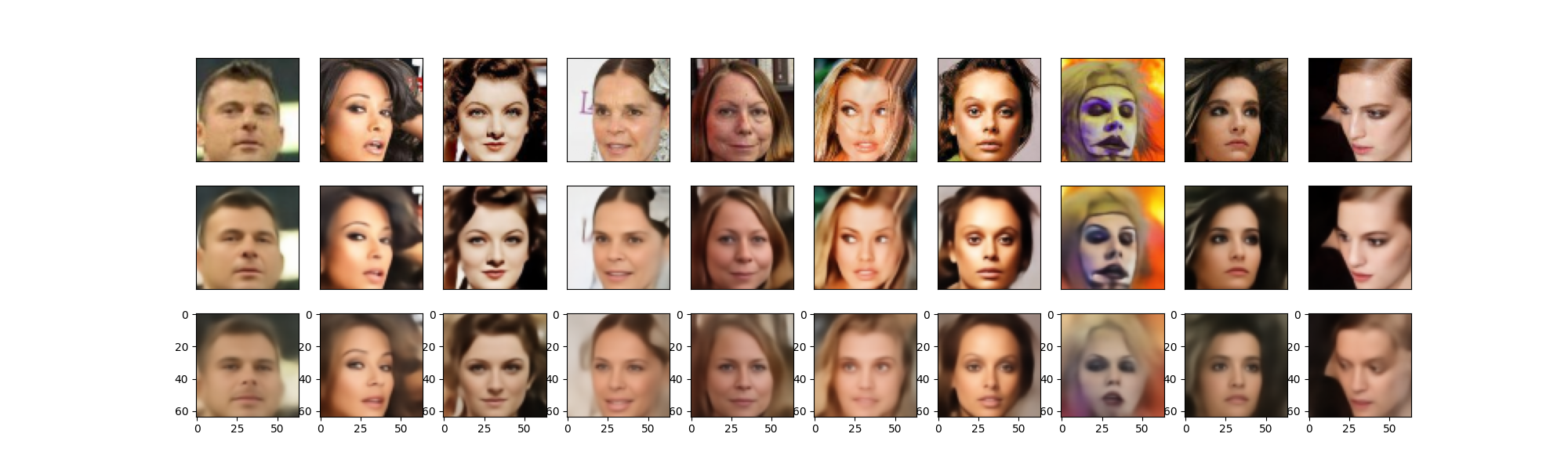}\vspace{-.2cm}
         \caption{Relocation of Type 1, between latent spaces relative to different training instances of the same generative model, in this case a particular Variational Autoencoder \cite{SVAE}. The two reconstructions are almost identical.}
         \label{fig:reloc_type_1}
     \end{subfigure}
     \hfill
     \begin{subfigure}[b]{\textwidth}
         \centering
         \includegraphics[width=\textwidth]{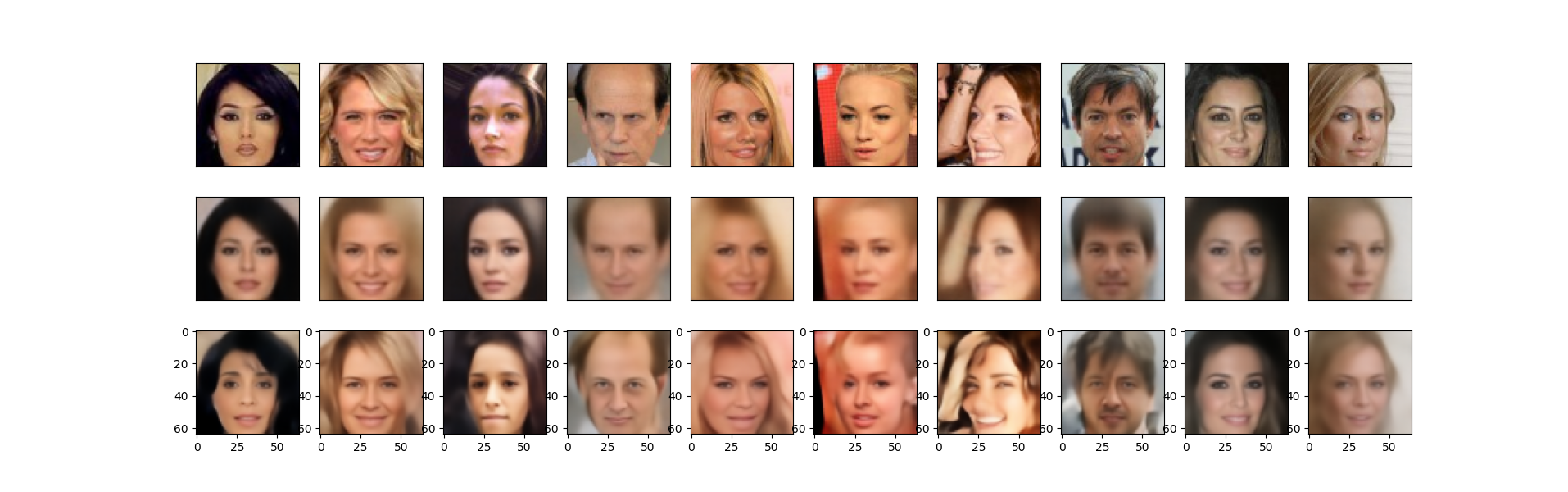}
         \caption{Relocation of Type 2, between a Vanilla VAE and a state-of-the-art 
         Split-VAE \cite{SVAE}. The SVAE produces better quality images, even if not necessarily in the direction of the original: the information lost by the VAE during encoding cannot be recovered by the SVAE, which instead makes a reasonable guess.}
         \label{fig:reloc_type_2}
     \end{subfigure}
     \hfill
     \begin{subfigure}[b]{\textwidth}
         \centering
         \includegraphics[width=\textwidth]{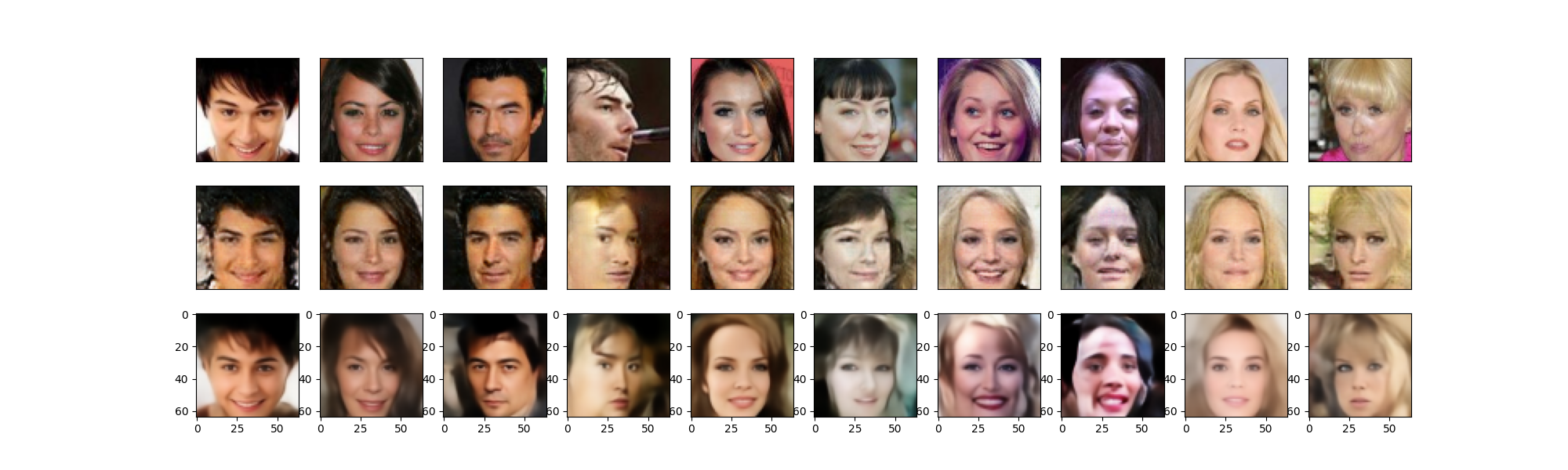}
         \caption{Relocation of Type 3, between a vanilla
          GAN and a SVAE. Additional examples involving StyleGAN are given in Section\ref{sec:StyleGANspace}. To map the original image 
(first row) into the latent space of the GAN we use an inversion network.  
Details of reconstructions may slightly differ, but colors pose and the overall 
appearance is surprisingly similar. In
some cases (e.g. the first picture) the reconstruction 
re-generated by the VAE (from the GAN encoding!) is closer to
the original than that of the GAN itself.}
         \label{fig:reloc_type_3}
     \end{subfigure}
        \caption{Examples of relocations of different Types. In the first row we have the original, in the second row the image reconstructed by the first generative model, and in the third row the image obtained by the second model after linear relocation in its spcae.}
        \label{fig:three}
\end{figure*}

The key questions we are interested in are the following:
\begin{enumerate}
    \item Do different trainings of the same generative model induce the extraction of similar
    features from data, and hence substantially isomorphic spaces up to, say permutations or linear
    transformations? We refer to this type of transformations as being of Type 1.
    \item Do different architectural models driven by common learning objectives (e.g. maximizing log-likelihood) learn similar
    features? How much do the extracted features depend on the neural network structure?
    We refer to this type of transformations, between spaces of variants of models in 
    the same class, as being of Type 2.
    \item Finally, what is the influence of the learning objective on the internal representation? 
    Is e.g. a Generative Adversarial Network learning the same features of a Variational AutoEncoder? We refer to these transformations as
    being of Type 3.
\end{enumerate}
Any answer, whether positive or negative, could substantially improve our knowledge of generative techniques.

Our surprising preliminary results, reported in this article, seem to suggest that (provided models have not been taught or explicitly conceived to act differently)
it seems to be possible to pass from a latent space to another by means of a {\em simple linear mapping} preserving most of the
information.

This linear transformation may be computed directly through linear regression, but we advocate a learning-based technique based on a suitable small ``support set" of data samples enucleating, in the visible space, the key variational factors of the data manifold. When we say ``small", we mean that the set has a cardinality comparable with the number 
of variables in the latent space (so, {\em really small}): for instance, in the case of CelebA, we experimented with a support set of 150 images. Locating these 150 samples in the two spaces is enough to allow the definition of a relocation map for all data.

The main results of our investigation are summarized in Figures~\ref{fig:three}. Figure~\ref{fig:reloc_type_1} describes an example of relocation between different trainings of a same network (relocation of Type 1); Figure~\ref{fig:reloc_type_2} is relative to the relocation between 
different models of a same class---two different VAEs, in this case (relocation of Type 2); Figure~\ref{fig:reloc_type_3} is an example of relocation from a VAE to a GAN, that is between different models with different learning objectives (relocation of Type 3). While details may slightly differ, especially
for transformations between different generative models, the
overall appearance (pose, colors and background) is substantially 
preserved. Considering the non-linearity of these generative processes, the result is, at a first glance, quite surprising: 
pairs of points related by a simple linear mapping in the latent spaces of two different generative models are decoded by the respective decoders in closely related---in some cases almost identical---images!

\subsection*{Structure of the article}
The structure of the article is the following. We start by providing, in Section \ref{sec:generative}, a quick introduction to generative modeling, and in particular
to latent variables models, comprising the popular Variational Autoencoders (VAEs) and Generative Adversarial Networks (GANs); in this section, we also discuss the problem of inverting GANs. Section \ref{sec:semantic_interpretation} cover the domain of
semantic exploration of latent spaces, representation learning and disentanglement.
In Section \ref{sec:datasets} we start introducing the datasets, the models and the methodology that we used for our experiments. Since we focus on linear transformations,
they can be defined by a small set of points, that we call Support Set: locating the 
points in the Support Set in the two latent spaces is enough to define the transformation. Our approach to get a good Support Set is discussed in 
Section \ref{sec:support_set}. In Section \ref{sec:results} we give numerical results about the mappings (visual examples, more readily interpretable, are spread over the article).
Section \ref{sec:StyleGANspace} is devoted to the discussion of the latent space of StyleGAN, 
that seems to present some pathological issues: many faces in the CelebA dataset lie
outside of its generative range. Even in this case, however, provided we confine the transformation to
the StyleGAN subspace, we discover interesting linear mapping to other spaces. 
Conclusion and future works are discussed in Section \ref{sec:Conclusions}. 
Additional material is given in appendix: 
a detailed description of the models used in this work
(Section \ref{appendix:models}), a full list of all images in the CelebA Support Set 
(Section \ref{appendix:labels}).

\section{Generative Modeling}
\label{sec:generative}
Generative modeling is the task of learning the high-dimensional probability distribution of a data manifold starting from a representative set of 
samples. When successfully trained, generative models can be used to create new samples from the underlying distribution, possibly providing estimations of their likelihood. 
The learning process provides an essential and valuable insight of the kind of features
used to encode the distribution, and the way the model ``interpreted'' and ``understood'' data.

At the heart of generative techniques there is a relatively small set of
techniques \cite{generative_introduction, generative_survey_2018}: Auto-Regressive models \cite{pixelrnn, pixelsnail}, Flow models \cite{kingma2018glow, dinh2016density, voleti2021multi, flow_methods}, Energy-based models \cite{ho2020denoising, du2019implicit, energy_models_survey} and Latent-Variable models, particularly GANs \cite{large_gan, InvidiaGAN18, Radford15} and VAEs \cite{Kingma13, vq_vae_2, IWAE}.

In this article, we shall mostly focus on the popular and effective Latent-Variable models, that is models
where the actual distribution $p(x)$ of a data point $x$ is expressed through marginalization
over a vector $z$ of {\em latent variables}:
\begin{equation*}
    p(x) = \int_{z} p(x\rvert z)p(z) dz = \EX_{p(z)} [p(x\rvert z)]
\end{equation*}
where $z$ is the latent encoding of $x$ distributed with a known distribution $p(z)$ named {\em prior distribution}. 
The distribution $p(x\rvert z)$ is usually learned by a deep neural network; after training
it can be used to generate new samples via ancestral sampling:

\begin{enumerate}
    \item sample $z \sim p(z)$;
    \item generate $x \sim p(x\rvert z)$.
\end{enumerate}



\subsection{Variational Autoencoders}
\label{sec:vae}

A Variational AutoEncoder (VAE) \cite{VAEKingma} has a structure similar to a classical auto-encoder \cite{DeepBook,autoencoders}, being composed of an {\em encoder} producing a latent vector $z$ from an input $x$ and of a {\em decoder} which reconstructs the input $\hat x$ from a latent code; the two components are simultaneously trained using, e.g., a mean squared error loss $\mid x - \hat x \mid_2$. However, in order to regularize the latent space, which is a precondition to support semantically meaningful generation \cite{generative_introduction}, latent variables are interpreted as parameters of a local distribution $q(z\rvert x)$ and a Kullback-Leibler component $KL(q(z\rvert x)\; \| \;\mathcal{N}(0, 1))$ is added to the reconstruction loss,  with the purpose of pushing the marginal distribution $q(z)$ towards a standard Gaussian $\mathcal{N}(0, 1)$.
Balancing of these two loss components, usually via a $\gamma$ or $\beta$ parameter, is crucial for better generation and learning of disentagled features \cite{vae_loss_balancing, beta-vae17, understanding-beta-vae18}.

Several issues affect the performance of VAEs, most importantly blurriness of generated images \cite{VAEGreen}. As such, many variants 
have been proposed over the years to improve results by addressing the mismatch between the aggregate inference distribution $q(z)$ and the prior $p(z))$. These comprise: quantization of the latent code (VQ-VAE \cite{VQVAE}), use of normalizing flows (Hybrid VAE \cite{autoregressive16}), two-Stage architectures \cite{2_stage_vae}, and hierarchical models \cite{DRAW,kingma2018glow}.


\subsection{Generative Adversarial Networks}
\label{sec:gan}
In a Generative Adversarial Network (GAN) \cite{GANs, tutorial-GAN, Radford15} a {\em generator}, acting as a sampler for the desired distribution, is jointly trained with a {\em discriminator}, evaluating  the output of the generator by attempting to distinguish real from generated  (``fake'') data. This can be formalized in the form of a zero-sum game, where one agent's gain is another agent's loss; the generator and the discriminator must be trained alternately, freezing the respective adversarial component; at the end of the process the generator is supposed to win, producing samples that the discriminator is unable to distinguish from real.

GANs are known to have unstable training and several issues among which the well known mode collapse phenomenon \cite{tutorial-GAN}. Indeed, multiple variations for the loss function have been studied over time \cite{GAN_losses_survey}, including the Wasserstein loss \cite{Wasserstein_GAN}, least squares loss \cite{LSL_GAN} or the introduction of a penalty term for the discriminator \cite{DRAGAN}. Furthermore, a myriad of variations on the structure itself have been proposed, among which: maximizing the mutual information between specific latent variables \cite{InfoGAN}; exploiting pairs of GANs to perform style transfer between images in distinct datasets \cite{cycleGAN}; GANs with attention layers \cite{attention_GAN}.

A particularly interesting series of works come from the application of style transfer concepts to GANs (StyleGAN and its successors \cite{karras2019style, karras2020analyzing, karras2021alias}). StyleGAN builds on Progressive GANs \cite{Progressive_GAN}, whose structure is unchanged from that of a baseline GAN but is implemented progressively: the architecture is trained starting from down-sampled images at very low resolution, and at each progression step the input size is increased while additional layers are introduced to both generator and discriminator.

\begin{figure}[ht]
\begin{center}
\includegraphics[width=\columnwidth]{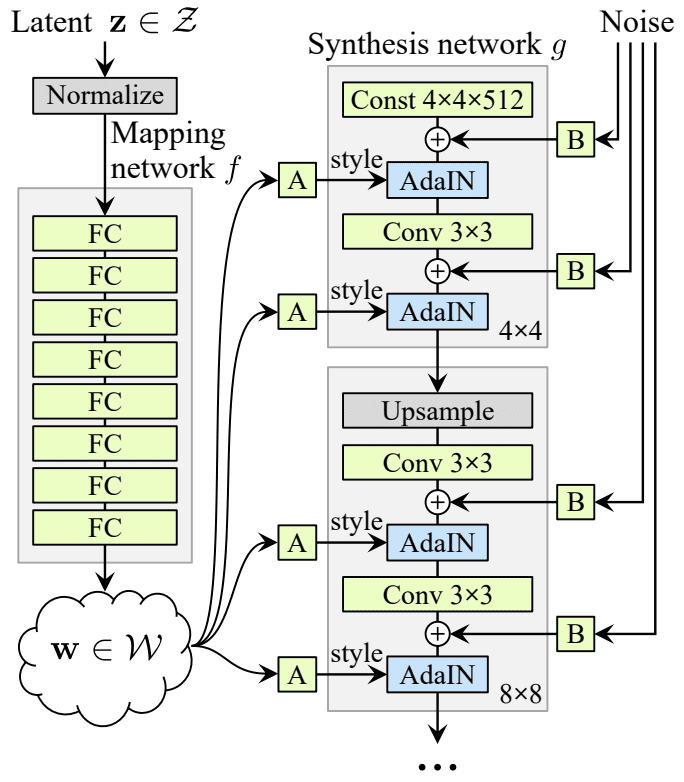}
\caption{Structure of the StyleGAN generative network (picture from \cite{karras2019style}).  Observe: (1) the two distinct latent spaces $Z$ and $W$;
(2) the {\em mapping network} taking a randomly sampled point $z\in Z$ as input and generating a {\em style vector} $w$; (3) the use of Adaptive Instance Normalizationation, or AdaIN (Blocks A), to apply style vectors after each convolution layer of the Synthesis network;
(4) the exploitation of noise as an additional source of randomness passed through learned scaling layers (Blocks B).
\label{fig:styleGAN}}
\end{center}
\end{figure}

StyleGAN further builds on this structure by adding to the generator (\textit{Synthesis} network) a fully connected \textit{Mapping} network which takes the usual seed $z \in Z$ and produces a ``style'' vector $w \in W$. This vector is then specialized per-layer through Adaptive Instance Normalization (AdaIN), which according to the authors produce a behavior similar to style-transfer. Furthermore, a small amount of noise is added to all blocks of the Synthesis network to better fill in the output details. The full structure of StyleGAN can be seen in Figure \ref{fig:styleGAN}.

\subsubsection{GAN Inversion}
\label{sec:inversion}
The generator of a GAN usually takes as input a seed $z \sim \mathcal{N}(0, 1)$, and has a role directly comparable to that of a VAE decoder. However, GANs lack a direct encoding process of the original input sample, unlike a VAE encoder. If, as is the case for our study, both generative and encoding processes are needed, a third neural network has to be added to a pre-trained GAN as a sort of plug-in encoder. This re-coder component is known as an inverse GAN, and building an accurate re-coder is a known problem in the literature \cite{gan_inversion_survey}.

\begin{figure*}[ht]
\begin{center}
\includegraphics[width=\textwidth]{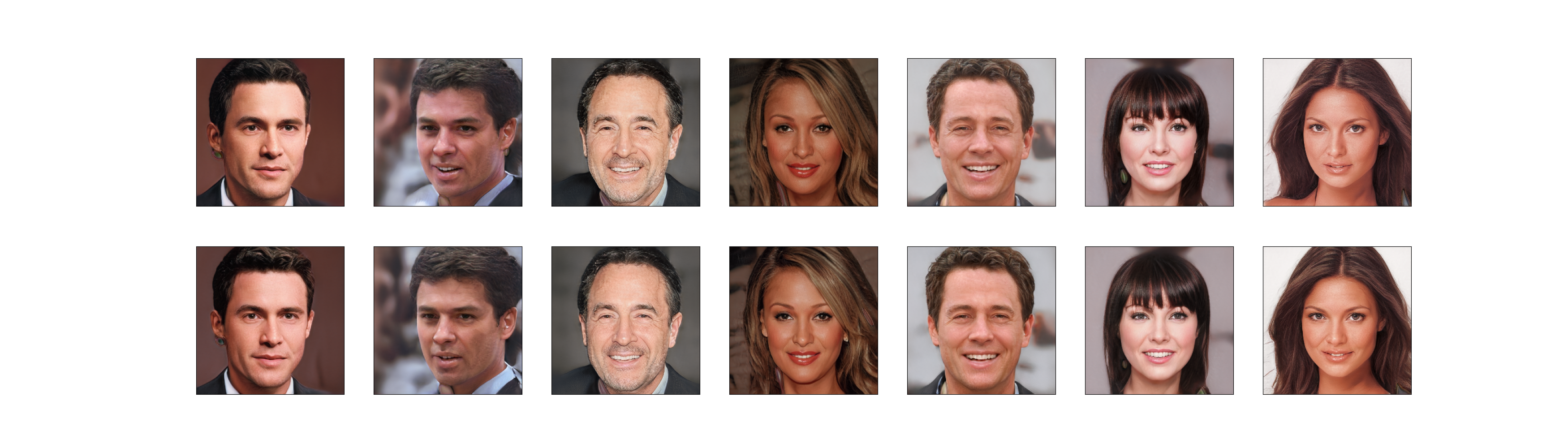}
\caption{Results of our own network for StyleGAN inversion. Images in the first row have been generated by StyleGAN; they are re-coded into the W space and regenerated (second row). 
The two images are hardly distinguishable. However, as we shall see in Section~\ref{sec:StyleGANspace}, inversion can be more problematic for images outside the generative range of the model; in principle, 
a good generative model should be able to produce any sample, provided it is 
not too atypical.\label{fig:gan_inv1}}
\end{center}
\end{figure*}

Several approaches to inversion have been explored \cite{perarnau2016invertible, bau2020semantic, daras2020your, anirudh2020mimicgan}, mostly for editing applications, the simplest being SGD optimization \cite{gan_inversion_optimization} or a learning-based approach such as using a neural network trained on generated images to reconstruct the original latent vector using a mean squared error loss $\mid z - \hat z \mid_2$, with the advantage that over-fitting is never an issue since training is not constrained to samples of the original data. Hybrid methods combining both efforts have also been explored \cite{gan_inversion_first_hybrid, gan_inversion_hybrid_2}.

Recent works have focused mostly on the inversion of the popular StyleGAN, building on previous work with a variety of inversion structures and minimization objectives \cite{stylegan_inversion, collins2020editing, abdal2020image2stylegan, stylegan_overparameterization, alaluf2022hyperstyle} with the aim of generalization to any dataset. However, we used a simpler and narrow approach by developing our own StyleGAN inverter for the $W$ space using a naive recoding network. It works surprisingly well for commonly generated samples, with a final mean square error close to $0.0040$. We show some examples of recoding in Figure~\ref{fig:gan_inv1}.

\section{Semantic Interpretation of Latent Spaces}
\label{sec:semantic_interpretation}
The latent space of a generative model efficiently synthesizes information from data, however the resulting compressed vectors cannot be easily mapped onto understandable features such as labels or attributes. Therefore, it is also unknown how exactly a model learns from data, in terms of how well it encodes its features, biases and human-meaningful characteristics. At the same time, this knowledge could fundamentally influence the quality of models and provide a foundation on which to improve their performances without relying solely on empirical and qualitative analyses.

Conditional architectures \cite{cVAE, InfoGAN} can indeed mitigate this issue by explicitly feeding features alongside samples during training, but in doing so they remodel the task as a supervised problem with respect to the classes on which conditioning is done, with all other data features remaining non-explainable. 
These approaches do not provide interesting information about the way the neural
network understand data, and for this reason, they will not be discussed in this work.

\subsection{Exploration and Disentaglement}
Many works attempt to understand the latent space of GANs by performing exploration on the latent space, that is, they introduce small nudges in a direction based on the empirical principle that they will correspond to a small change in the corresponding generated data. The approach can be particularly useful for image editing, as once a semantically meaningful direction is found (eg. color, pose, shape), it can be traveled to tweak an image, introducing a desired feature without the need for a conditional generation model. InterFaceGAN \cite{Shen22} supposes that for a given feature taking values in $(-\infty; \infty)$ there exists an hyperplane in the latent space whose normal vector allows for a gradual modification of the feature, which can be found e.g. via an SVM \cite{SVM}. Further work based on this idea searches for these directions as an iterative or an optimization problem \cite{li2021interpreting} and also extend it to controllable walks in the latent space \cite{li2021discovering}.

A different, more systemic approach to the problem is by \cite{Closed-Form-fact}, which use a closed-form equation to find the editing direction $n_i$ applied per-layer $i$ of a generator, which is then composed to find the overall direction $n$. Another approach of the same ``arithmetic" flavor comes from \cite{Exploration_PCA}, where a generative application of PCA with a non-linear kernel is used to determine the hidden features of a small-scale dataset, without any reliance on a particular generative model.

Much less work on exploration has been devoted to VAEs. An example is given by \cite{learning_latent18}, which however works on a conditional architecture, in order to produce lower-dimensionality subspaces that are easier to analyze.




\section{Datasets, Models, Methodology}
\label{sec:datasets}
\subsection{Datasets}
As stated in the abstract, we confined our analysis to the familiar and largely investigated data manifold of human faces.
Our dataset of reference is CelebA \cite{celeba}, including its higher-quality version CelebAHQ \cite{InvidiaGAN18}. Images taken from CelebA have been aligned as per their paper \cite{celeba} and then cropped to size $128\times 128$ with a $y$ offset of $45$ and an $x$ offset of $25$ in order to remove as much background information as possible. The crop is then downsampled to size $64\times 64$ with bilinear interpolation).

CelebaHQ is a dataset of 30K images at resolution 
$1024\times 1024$, obtained from a subset of CelebA with a
complex methodology explained in appendix C of \cite{Progressive_GAN},
comprising a sophisticated preprocessing phase, super-resolution
techniques, and selection of best quality samples. 


\subsection{Generative models}
\label{subsec:models}
For our experiments we took into considerations 4 different models, two GANs and
two VAEs; in each class, we investigated a basic, average quality "vanilla" version and a more sophisticated, state-of-the-art model. A summarizing Table \ref{table:models_basic} for these models is provided. More in-detail, we have investigated the following architectures:
\begin{enumerate}
    \item Vanilla VAE \cite{VAEKingma} using $\gamma$ balancing \cite{vae_loss_balancing} with a latent dimension $Z$ = 64 trained on the cropped CelebA;
    \item Vanilla GAN \cite{GANs} with a latent dimension $Z$ = 64 trained on the cropped CelebA;
    \item SVAE \cite{SVAE} with a latent dimension $Z$ = 150 trained on the cropped CelebA;
    \item StyleGAN \cite{karras2019style} pre-trained on CelebA-HQ, which has a latent dimension $Z$ of size $512$ and a style-vector latent dimension $W$ of the same size.
\end{enumerate}
The structure of the StyleGAN has been already briefly discussed in Section \ref{sec:gan}. The in-depth architecture of the other models, not central to the topic of this article, is given in Appendix \ref{appendix:models}. 


The dimension of the latent space and the resolution of 
the different models is summarized in Table \ref{tab:modelinfo}.

\begin{center}
\begin{table}
\centering
\begin{tabular}{ |c|c c|} 
 \hline
 \textbf{Model} & \textbf{Latent dim} & \textbf{Resolution} \\ 
 \hline
 GAN & 64 & $64 \times 64$ \\ 
 VAE & 64 & $64 \times 64$  \\
 SVAE & 150 & $64 \times 64$ \\\hline  
 StyleGAN & 512 & $1024 \times 1024$ \\ 
 \hline
\end{tabular}
\caption{\label{table:models_basic} Dimension of the Latent Space and Resolution for
the different models. 
\label{tab:modelinfo}}
\end{table}
\end{center}

\subsection{Methodology}
For each one of the previous models, apart StyleGAN where we only had at our disposal a single set of pre-trained parameters, we trained and tested five different instances.
When reporting values in the results, if not differently stated, they have to be
understood as an average over the different trainings.

Mapping between different models (transformations of Type 2 and 3) can have a lot of additional issues. Firstly, the two latent spaces may have sensibly different dimensions, for instance $512$ for StyleGAN versus $150$ for the SVAE and  for the other models, and may work at different resolutions, for instance $1024\times 1024$ for StyleGAN versus $64\times 64$ for the other models. Furthermore, the two generative models may have been trained on the two different datasets which, albeit similar, have different data
and different crops. To this aim, when
passing from CelebA-HQ to CelebA we take a simplified crop of dimension $880\times 880$ with an
height offset of 20 and a width offset of 60, which is then downsampled to size $64\times 64$ with bilinear interpolation.

Since we are interested in linear mappings, the transformations may be defined by 
a small set of "corresponding" points common to both spaces: this is what we call a Support Set.
Our methodology to build it is defined in Section \ref{sec:support_set}.
The support Set is defined in the visible domain; we trace their respective encodings
in the different spaces, and define the map by linear regression with 
mean squared error as a loss. When we cannot use a Support Set, we may directly work
with the whole visible domain (or the subset of the visible domain common to the two
spaces), sampling minibatches in it.

\section{Support Set}
\label{sec:support_set}
In this Section we explain the technique used to build a 
small support set of examples driving the linear transformation.
This is based on the following steps, each one detailed in a respective subsection: 
\begin{description} 
\item[features ordering] \hspace{1mm} we order latent 
variables according to their relevance for reconstruction, using a suitable metric discussed below;
\item[features selection] \hspace{1mm} we select  a small number $n$ of particularly significant latent variables; $2^n$ must be lower
than the cardinality of the support set;
\item[sample selection] \hspace{1mm} we select points in the space belonging
to extremal regions with respect to the selected features.
\end{description}

\subsection{Features ordering}
\label{sec:features_ordering}
Feature importance---the task of associating 
a score to input features based on how useful they are for
solving a specific problem---is a major subfield of Machine
Learning. In the case of generative modeling, the goal is
to maximize the (log)likelihood of data, and it is natural
to associate a score to features according to their 
contribution to this objective. It is worth observing that
different techniques, like e.g. PCA, would not be beneficial 
to this aid, due to the shape of the prior latent 
distribution which is, typically, a spherical Gaussian 
distribution\footnote{Even the potential mismatch between the prior and
the aggregate inference distribution in the case of VAEs 
cannot be exploited by PCA, since this technique only takes
into consideration the first two moments of the distribution.}.

Our feature importance technique requires an encoder in addition to a decoder: it fits particularly well with VAEs, but 
it can be generalized to GANs by exploiting a re-coder network (see Section~\ref{sec:inversion}). 
Specifically, in order to evaluate the contribution of the variable to the loss function, we compute over a large number of data the average difference between the reconstruction error when the latent
variable is zero-ed out with respect to the case when
it is normally taken into account. 
We call this information the {\em reconstruction gain} 
associated with the latent variable. It was introduced in
\cite{sparsity} where it was used to compare the reconstruction error and 
the Kullback-Leibler divergence on a per-variable base, in order to clarify the variable collapse phenomenon  \cite{BurdaGS15,RobustPCA,PosteriorCollapse}.

\begin{figure}[ht]
\begin{center}
\includegraphics[width=\columnwidth]{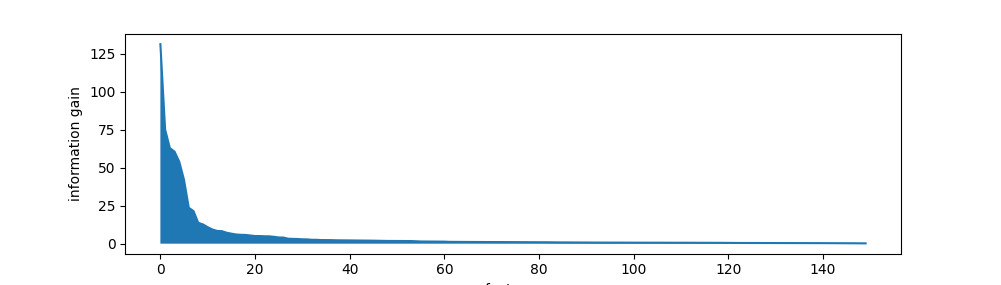}
\caption{Information gain for all variables, in
decreasing order. Only a bunch of variables are in charge
of the macroscopic factors of variations.
\label{fig:info_gain}}
\end{center}
\end{figure}

We did the experiment on the SVAE, which in our experiments has a latent space of 150 variables. In Figure~\ref{fig:info_gain} we show the information gain relative to all its latent variables, ordered by relevance.

Eleven variables have a score higher than 10, although the distribution has a relatively long tail: the first 20 variables are responsible for about 75\% of the information.

\begin{figure*}[ht]
\begin{center}
\includegraphics[width=\textwidth]{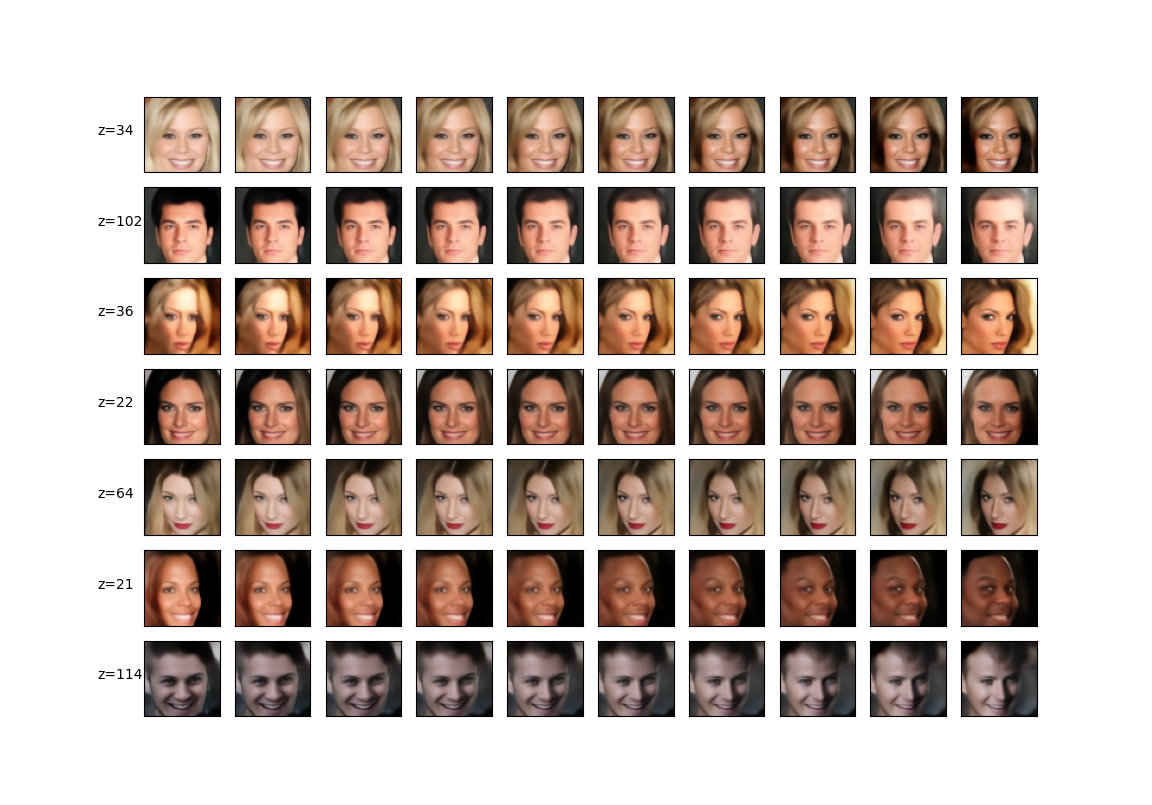}\vspace{-0.8cm}
\caption{Effect of the seven most informative latent variables
in the visible domain. Each image is obtained by varying a
specific variable in the range [-2.25; +2.25]. 
Considering these are the variables with the largest information
gain, it may be argued that their impact is less pronounced than expected. 
Most of the variables are associated with a change in luminosity
of all or part of the image, possibly associated with modifications in hair color, source of illumination
and tiny variations in the pose. In the case of variable 21, 
there seems to be progressive Female-Male transition (and vice-versa for variable 114).
\label{fig:latent_vars}}
\end{center}
\end{figure*}

\subsection{Feature Selection}
We keep a small number of the most informative variables.
For the way we shall use it, this number must be smaller 
than the logarithm of the cardinality of the support set.
In our case, we aim to a support set of dimension 150, 
so we focus on the 7 most relevant variables. 

In Figure~\ref{fig:latent_vars} we show examples of the effect of some of these variables on generated images: we take a random point and progressively modify the
given variable in the range between -2.25 and 2.25 (remember that 
the latent space standard deviation is 1). 

\subsection{Sample selection}\label{ss:sample_selection}
Finally, we divide the latent space in sectors corresponding
to extreme values for the previously selected variables, and
pick up samples in these sectors. 
\begin{figure}[ht]
\begin{center}
\vspace{-.5cm}\includegraphics[width=\columnwidth]{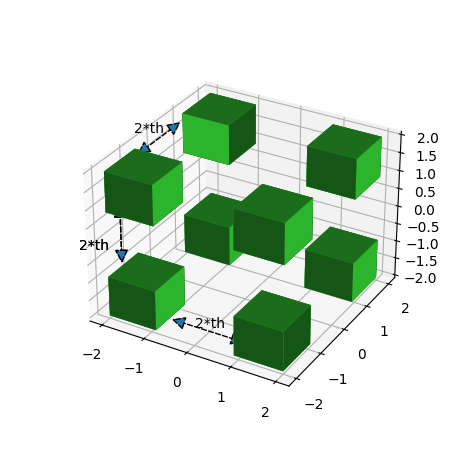}\vspace{-0.8cm}
\caption{Example of sectors in 3 dimensions (cropped to distance 2
from the origin). The distance between sectors is equal to twice a
configurable threshold. We work with the 7 most informative latent
variables, obtaining a total of $2^7=128$ sectors.
\label{fig:sectors}}
\end{center}
\end{figure}

More precisely, having defined a threshold {\em th} and a "direction" {\em dir} given by a $+/-$ sign for each selected variable,
a sector defined by the pair $(th,dir)$ is the set of points with
direction compatible with {\em dir} and at a distance from the origin
larger than {\em th}. Since we consider all possible directions,
this gives a total of $2^n$ sectors where $n$ is the number of selected variables (for a fixed {\em th}).
In each sector, we pick up a sample at random (enlarged {\em th} sectors become progressively less inhabitated). 

It is interesting to observe that the number of latent points
in the dataset within different sectors at a given 
threshold is far from uniform. This seems to be a confirmation 
that the actual image distribution is far from the desired Gaussian
normal prior and, in a VAE, a symptom of the potential mismatch between the generative prior and the aggregate inference distribution computed by the encoder, which is a well known
and problematic aspect of VAEs 
 \cite{ELBOsurgery,rosca2018distribution,aboutVAE}. 
Attempts to solve this issue have been
made both by acting on the loss function \cite{WAE} or by exploiting more complex priors \cite{autoregressive16,Vamp,resampledPriors}; the actual effects on the latent space of these techniques is an interesting 
research direction for future investigations.

In Figure we show typical inhabitants for a few given sectors. As expected, 
they share macroscopic features like background color, pose, hairs, and illumination.

\begin{figure}[ht]
\begin{tabular}{c}
\includegraphics[width=\columnwidth]{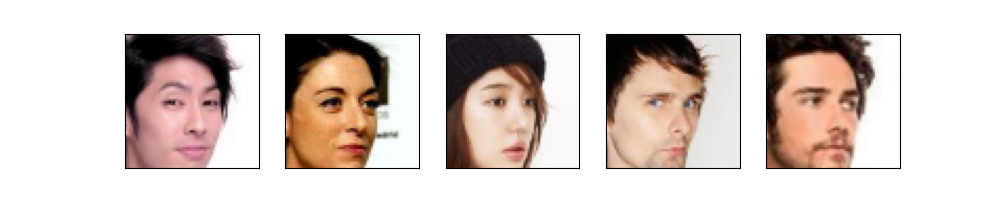}\vspace{-.4cm}\\
\tiny{sector 102}\vspace{-.2cm}\\
\includegraphics[width=\columnwidth]{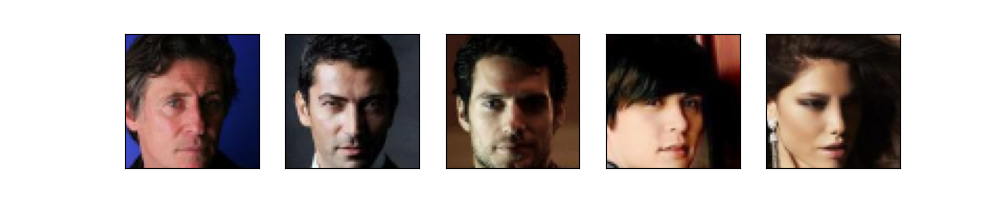}\vspace{-.4cm}\\
\tiny{sector 109}\vspace{-.2cm}\\
\includegraphics[width=\columnwidth]{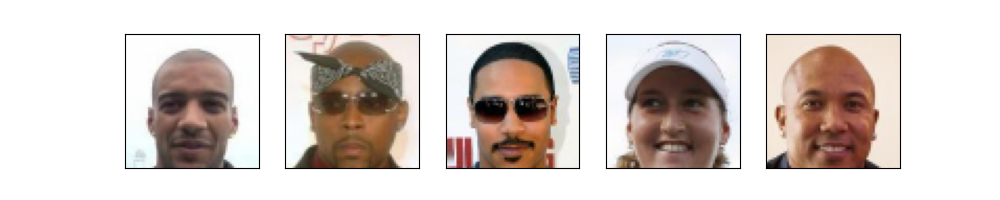}\vspace{-.4cm}\\
\tiny{sector 126}\vspace{-.2cm}\\
\includegraphics[width=\columnwidth]{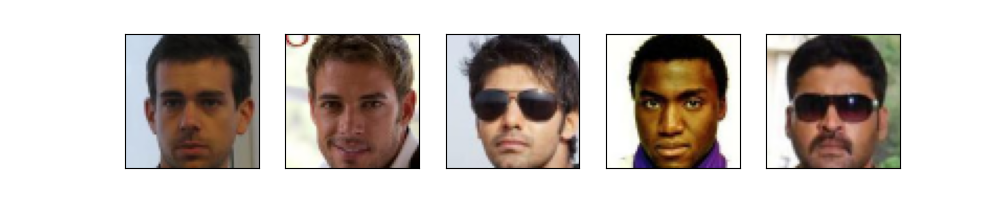}\vspace{-.4cm}\\
\tiny{sector 60}
\end{tabular}
\caption{Examples of data in different sectors. For each sector, 
images are different, 
but share macroscopic features: background color, pose, hairs, illumination,
etc.
\label{fig:samples_in_sectors}}
\end{figure}

Part of the 128 images resulting from our selection process are depicted in 
Figure~\ref{fig:support_set1}. The complete list of labels for the support set 
is reported in the appendix. 
\begin{figure}[ht]
\includegraphics[width=\columnwidth]{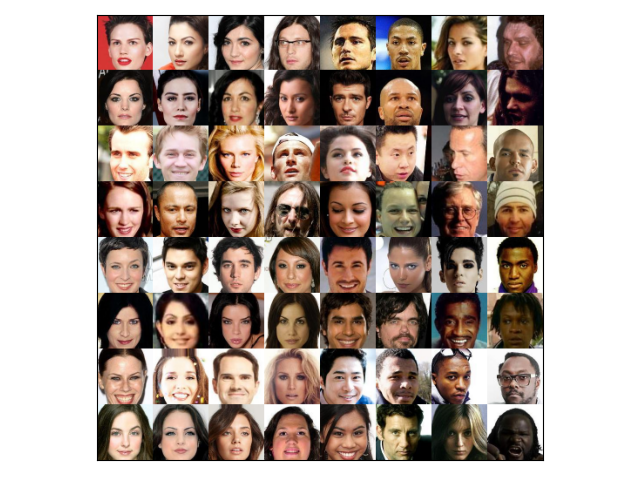}
\caption{Part of the images in the support set resulting from our 
selection process. The samples are supposedly representative of the
principal factors of variations in the dataset. Additional examples are given in
the appendix.
\label{fig:support_set1}}
\end{figure}
The samples in the support set occupy ``extreme" 
positions in the latent space with respect to the most informative 
directions: for this reason, they as supposed to be representative of the
principal factors of variations in the dataset.

As a partial confirmation of the previous hypothesis, we expect the distance 
between elements in the support set to be sensibly higher than the average
distance between points in the full dataset. This is actually the case: 
the mean squared error between random CelebA images is 0.116, versus 0.183 for
samples in the support set.

\section{Results}
\label{sec:results}
This Section contains numerical results relative to the transformation between latent spaces. 
The discussion of StyleGAN, for its relevance and some
interesting pathological issues, will be postponed to the next Section.

Here, with we shall use the names VAE, GAN and SVAE to refer to our specific implementations
of these models, discussed in Section \ref{subsec:models} and detailed in appendix \ref{appendix:models}.

We build a set of correspondent input-output pairs by encoding the Support Set (or the full 
set of visible data) into the two latent spaces. Then, we directly build a linear map by linear regression, minimizing the mean squared error between target and computed latent vectors.

\begin{figure}[ht]
\begin{center}
\includegraphics[width=.9\columnwidth]{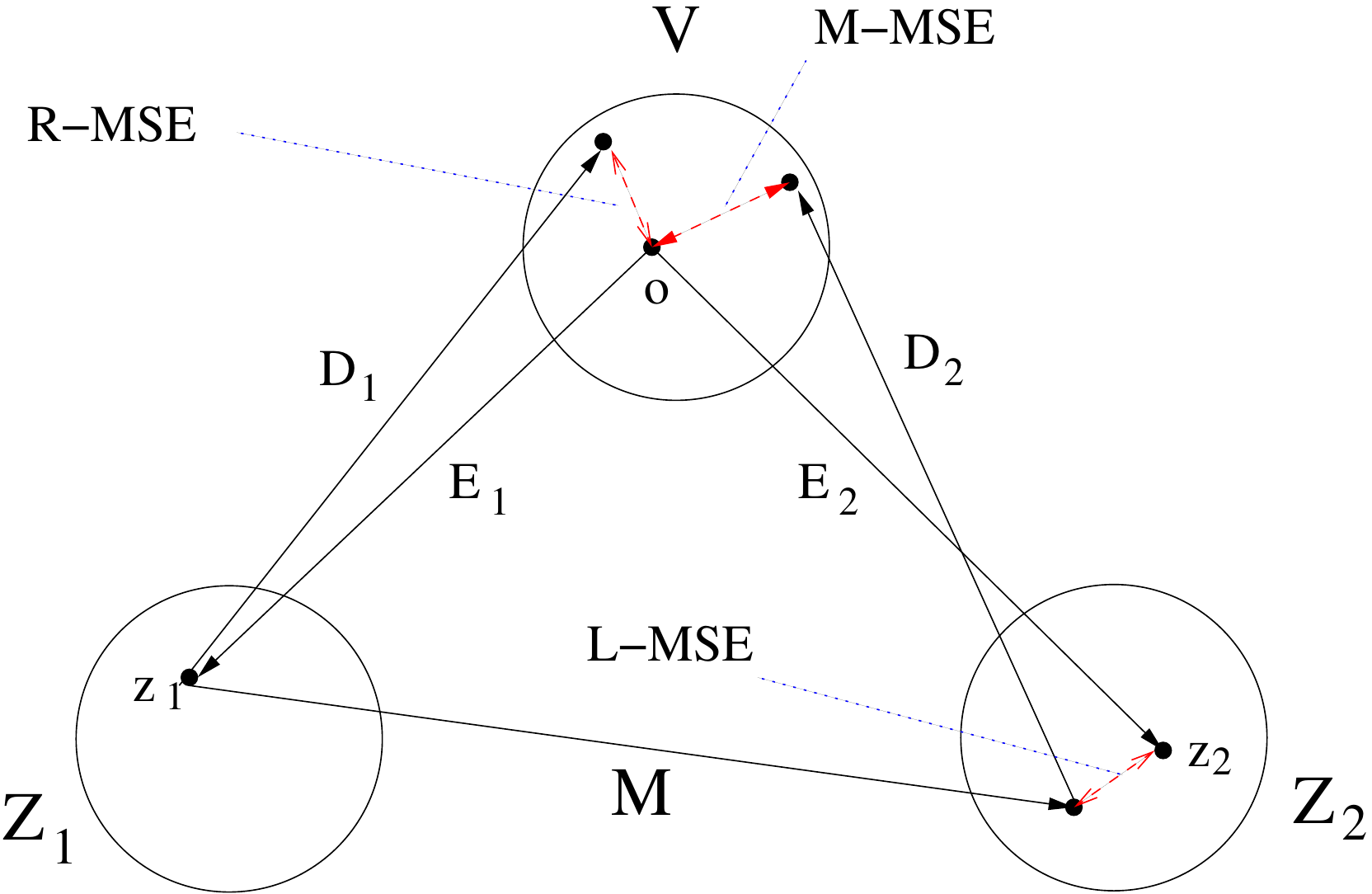}
\end{center}
\caption{Relocations Errors. An original point $o$ in the visible domain is mapped into internal representations $z_1$ and $z_2$ in the latent spaces $Z_1$ and $Z_2$. The map $M$ is trained to reconstruct $z_2$ from $z_1$: L-MSE is the mean squared error between $z_2$ and $M(z_1)$. R-MSE is the mean squared error, in the visible domain, between $o$ and its reconstruction according to the first generative model. M-MSE is the mean squared error, in the visible domain, between $o$ and $D_2(M(z_1))$. 
\label{fig:errors}}
\end{figure}

For each transformation, we provide three values:
\begin{description}
\item[L-MSE] Latent Mean Squared Error. This is the loss of the model, namely the mean squared error between the target vectors and those computed by the model;
\item[R-MSE] Reconstruction Error. This is the mean squared error between the original image in the visible domain and its reconstruction via the source generative model;
\item[M-MSE] Mapped Error. This is the mean squared error, in the visible domain, between original images and images reconstructed by the target generative model after linear mapping.
\end{description}
The three errors are graphically described in Figure \ref{fig:errors}.

The latent error L-MSE is not easily deciphered; the comparison between R-MSE and M-MSE provides a more intelligible information about the quality of the translation. 

Results are given in Table \ref{tab:results}.

\begin{table}
\centering
\begin{tabular}{ |c c|c c c| }
 \hline
 \textbf{From} & \textbf{To} & \textbf{L-MSE} & \textbf{R-MSE} & \textbf{M-MSE} \\\hline
 VAE & VAE & 0.03 & 0.0073 & 0.0103\\
 VAE & SVAE & 0.72 & 0.0073 & 0.0105\\
 VAE & GAN & 0.49 & 0.0073 & 0.0339\\
 GAN & VAE & 0.50 & 0.0284 & 0.0254\\
 GAN & SVAE & 0.86 & 0.0284 & 0.0275\\
 GAN & GAN & 0.43 & 0.0284 & 0.0335\\
 SVAE & VAE & 0.195 & 0.0035 & 0.0125 \\
 SVAE & GAN & 0.63 & 0.0035 & 0.0388 \\
 SVAE & SVAE & 0.20 & 0.0035 & 0.0067 \\
 \hline
 \hline
\end{tabular}
\caption{\label{table:mapping_results}Mapping results for different model pairs: (L-MSE) MSE between the target and Mapped Latent vectors; (R-MSE) MSE between the original and Reconstructed (encoded-decoded) images; (M-MSE) MSE between the original and mapped images via the learned linear mapping. When source and target coincide, we mean different trainings of the same model (Type 1 transformations).}
\label{tab:results}
\end{table}
For the sake of comparison, it it worth to recall that the mean squared error between CelebA images
is $0.116$; in all models the M-MSE is always below $0.039$. 

\section{The StyleGAN space}
\label{sec:StyleGANspace}
The ``extreme'' nature of the images in the Support Set makes them a very natural benchmark of the expressiveness of generative models: is it possible to reconstruct these images by passing them through an encoding-decoding process?

\begin{figure*}[ht]
\includegraphics[width=\textwidth]{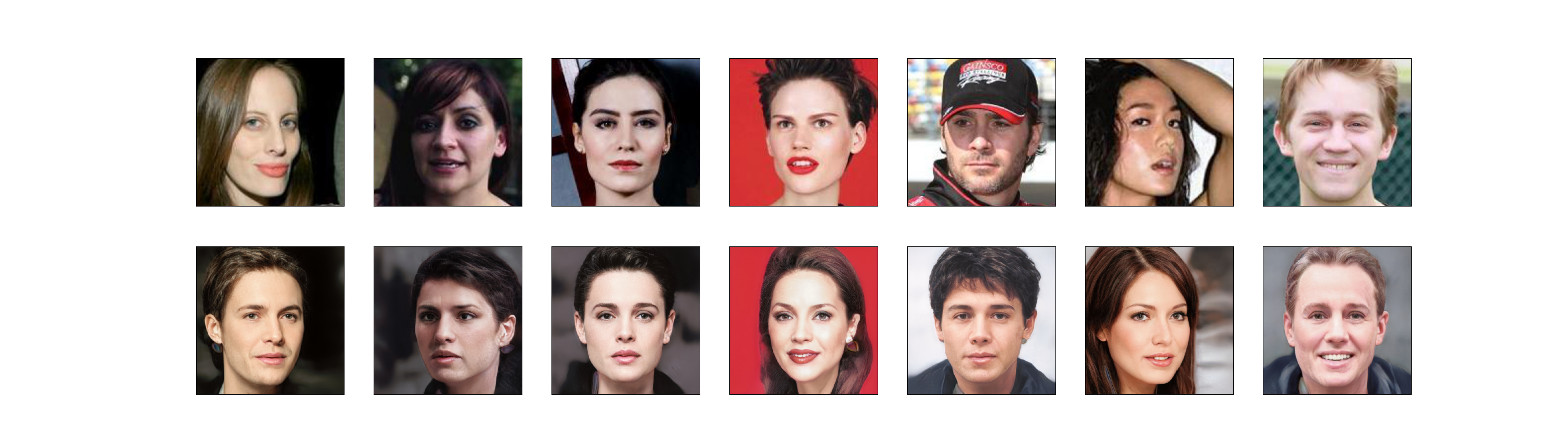}
\caption{StyleGAN inversion on images in the Support Set.
The macro structure (background, pose, illumination, etc.)
is preserved, but all other features are lost: images
in the Support Set seem to lie outside of the generative
range of StyleGAN. Note also the more ``conventional'' nature
of the images obtained by the inversion.
\label{fig:support_gan}}
\end{figure*}

For StyleGAN trained on CelebA-HQ, results are disappointing (see Figure \ref{fig:support_gan}, and compare them with the inversion of generated images in Figure~\ref{fig:gan_inv1}). Although the macrostructure is preserved (background, pose, illumination), details are sensibly different. Numerically, 
while the average mean squared error on generated images is $0.026$, the corresponding value for the Support Set is  $0.251$, almost ten times higher.

\begin{figure}[ht]
\begin{center}
\includegraphics[width=.3\columnwidth]{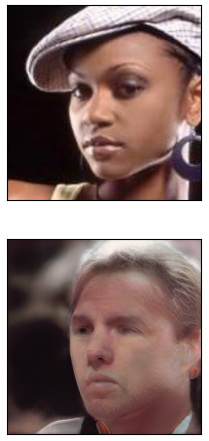}
\includegraphics[width=.3\columnwidth]{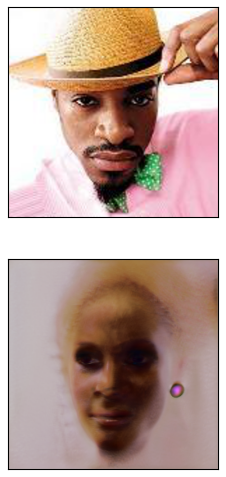}
\includegraphics[width=.3\columnwidth]{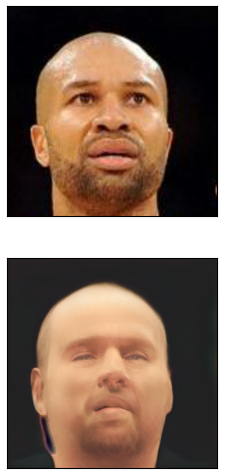}
\caption{Gradient ascent technique for StyleGAN on data
in the Support Set. The original is in the first row, and the
image generated through gradient ascent, in the second. The technique confirms that these images cannot be generated by StyleGAN.
\label{fig:gradient_ascent}}
\end{center}
\end{figure}

Our conjecture is that StyleGAN is simply unable to generate data in the support set: they do not belong to its latent space, specifically due to its training dataset. To check this claim we implemented a gradient ascent technique to generate latent representations corresponding to a desired output. Once again, the gradient ascent technique provides almost
perfect results on generated images but substantially fails
on images in the CelebA support set, as shown in Figure \ref{fig:gradient_ascent}.

\begin{figure}[ht]
\begin{center}
\includegraphics[width=\columnwidth]{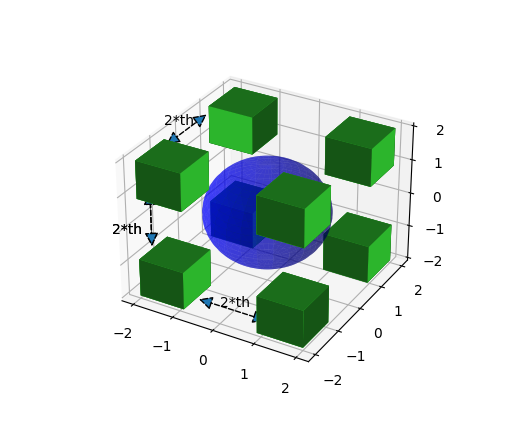}
\caption{CelebA Sectors seem to be external to the latent space of StyleGAN
\label{space_vs_sectors}}
\end{center}
\end{figure}

We believe that the latent space of StyleGAN, trained on CelebA-HQ, only faithfully reflects a subspace of the latent space of our other models, trained on the full CelebA 
dataset. In particular, points in our extreme {\em sectors} seem to lie
outside of the generative range of StyleGAN, or to be 
severely underrepresented (Figure~\ref{space_vs_sectors}).
The problem is possibly also related to the well-known fact that faces generated by StyleGAN (and other generative networks) can be easily distinguished from reals \cite{Distinguishing2022,deep_fake_recognition20,attributing19}.

\subsection{Comparison with different spaces} 
Since exploiting the Support Set is not a viable solution, we need to define a direct
mapping by regression on all data. Furthermore, we choose to work with the $W$ StyleGAN space, since the $Z$ space is passed through a series of fully connected layers (the {\em Mapping} network) which we suppose, by construction, cannot be inverted linearly.
Here we try to map the $W$ space of StyleGAN, trained 
over CelebA-HQ, to the latent space of SVAE trained over CelebA. 
The input to the transformation map is the vector $w$, obtained by ancestral sampling from the
$Z$ space. The expected output $z$ is obtained by synthesizing with StyleGAN the image 
corresponding to $w$, cropping and resizing it to dimension $64\times 64$ and encoding it 
in the SVAE latent space. The result of the linear map will be called $\hat{z}$; let 
$\mathit{SVAE}(z)$, and $\mathit{SVAE}(\hat{z})$ the corresponding decodings to the visible domain. 
As usual, input vectors $w$ may be generated {\em ad libitum}, with no risk of overfitting.

After training, the mean squared error between $z$ and $\hat{z}$ is around $0.45$ 
with a standard deviation of $0.05$. 
The mean squared error between $\mathit{SVAE}(z)$, and $\mathit{SVAE}(\hat{z})$ is $0.014$ 
with standard deviation of $0.002$.
All results have been repeated over 5 different parameters configurations of SVAE, relative to
5 different trainings (obviously, each experiment results in a different linear transformation).

\begin{figure*}[ht]
\includegraphics[width=\textwidth]{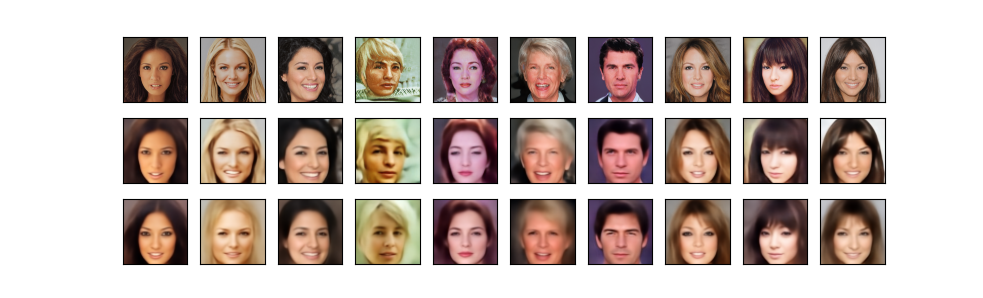}
\caption{Mapping from the W space of StyleGAN to the latent space of SVAE. 
In the first row we have sources, sampled by StyleGAN from $w\in W$. 
In the second row we have the
SVAE reconstruction, starting from a suitably cropped and rescaled images 
(SVAE work at resolution $64\time 64$): these images are the best possible approximation
of the source images obtainable by SVAE. In the third row we show the output produced
by the SVAE decoder after mapping each $w$ in its latent space: results are very similar
to those of the second row.
\label{fig:StyleGAN2SVAE}}
\end{figure*}

Result are shown in Figure~\ref{fig:StyleGAN2SVAE}. They are not perfect, but definitely interesting.

\begin{figure*}[ht]
\includegraphics[width=\textwidth]{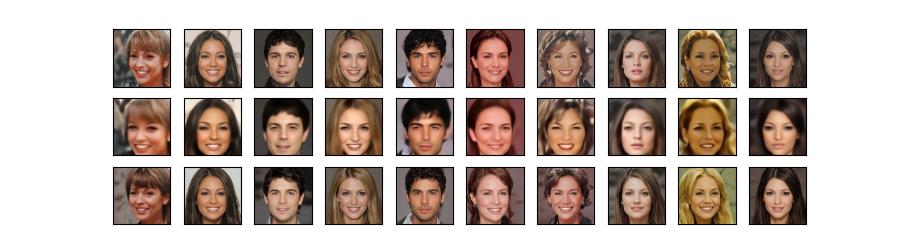}
\caption{Mapping from the latent space of SVAE to the W space of StyleGAN. 
In the first row we have images generated by StyleGAN: $StyleGAN(w)$, for $w\in W$.
In the second row we have their
SVAE reconstructions, starting from suitably cropped and rescaled versions. 
Images in the third row are obtained by first encoding $StyleGAN(w)$ in the latent space
of the SVAE, obtaining a latent representation $z$. This $z$ is then linearly transformed 
to a vector $\hat{w} \in W$; the final image is $StyleGAN(\hat{w})$.
\label{fig:VAE_to_styleGAN}}
\end{figure*}

We also tested a few variants weighting the distance between latent variables according to
their ``information relevance", but we did not observe significant improvements.

Let us come to the mapping from the latent space of VAE to that of the StyleGAN. To train
the transformation model (as usual, a single dense layer with no bias), we simply invert input and 
output of the previous network. After training, the mean squared error between $w$ and $\hat w$ 
is around $0.029$ with a standard deviation of $0.004$. 
The mean squared error between $StyleGAN(w)$, and $StyleGAN(\hat{w})$ is $0.076$ 
with standard deviation of $0.014$.
Results are visually really good, as can be visually checked in Figure~\ref{fig:VAE_to_styleGAN}. 

\section{Conclusions}\label{sec:Conclusions}
In this article we addressed the problem of comparing the latent space of different generative models, defining transformations
between them. Specifically, we proved that we can pass from a latent space to another by means of a simple {\em linear map} preserving most of the information. Hence, the organization of the 
latent space seems to be largely independent from
  \begin{itemize}
  \item the training process
  \item the network architecture
  \item the learning objective: GANs and VAEs share the same space
  \end{itemize}
The result is original, surprising and largely unexpected; apparently, the latent space, if not 
artificially constrained with different objectives, seems
to naturally organize itself in a way that is merely dependent
from the data manifold. Of course, we expect that this ``natural" structure
can be altered in many different ways, e.g. through conditioning, 
which strongly impacts the latent structure, or via transformations 
like normalizing flows, explicitly aiming towards a strong regularization 
of the space. We also do not expect the two spaces $Z$ and $W$ of StyleGAN to be linearly related, since otherwise
the long chain of 8 dense layers between them would have no purpose.

Our result is full of implications from the point of view of representation learning and disentanglement. The fact that the latent space has a sort
of implicit and native structure raises promising expectations about the
possibility of learning features in a completely unsupervised 
way. Moreover, the recent observation \cite{Shen22, li2021interpreting} that variations over a single semantical feature is a quasi-linear manifold in the latent space of
generative models fits well with our empirical observations, 
opening interesting perspectives about the possibility of
``porting'' disentanglement between different spaces, and 
more generally, to better understand the issue in a more general
framework.

The fact that the transformation between spaces is linear
obviously permits its definition in terms of a small set of
independent points of the same cardinality of the dimension
of the latent space; this is what we call a {\em Support Set}.
Locating these points in the two latent spaces is enough to
define the map.
In principle, any set of independent points could serve as a 
Support Set, but for robustness reasons, it seems preferable
to chose points as apart as possible between each other. 
We described a possible approach for defining such a set,
based on "sectors" in the space. This set is of interest in
its own, as it is representative of the
principal factors of variations in the dataset. Due to 
this fact, it also provides a natural benchmark
to test the expressiveness of generative models. 

This leads to an additional side contribution of our work:
in contrast with the usual belief, StyleGAN trained on CelebA-HQ seems to have serious generative deficiencies: many images, in particular most of the images in our Support Set from CelebA, seem to lie outside the generative range of StyleGAN. In particular, as it is also evident in inversion results, The StyleGAN generative process is privileging standardization, strongly penalizing defects, oddities and eccentricities: the StyleGAN space is not a space for minorities.

This could be a cause for concern about CelebA-HQ. Not only
it is computationally demanding, but one could also wander if it has statistical relevance: an assortment of 30K images in a space of dimension $3\times2^{20}$ looks more like a collection of scattered points than a data manifold. 

Our results also raise serious worries  about the increasing
use of generative techniques for data augmentation purposes.
All generative techniques seem to have serious biases, 
privileging likelihood over diversity: using them for data
augmentation may have no statististical significance. It is a bad practice that should be discouraged and deprecated.

As for future developments, most of the work just lies ahead. Here is a short, not-exhaustive list of possible topics:
\begin{itemize}
\item test and hopefully confirm our mapping results on different datasets;
\item deepen the relationship between the field of disentanglement through suitable linear manipulations of the latent space;
\item define and test a Support Set for StyleGAN and Celeba-HQ;
\item investigate the possibility to improve the transformation with residual non-linearities, and in that case study them;
\item better investigate and possibly find a remedy to the generative deficiencies of StyleGAN.
\end{itemize}


\subsection*{Data Availability}\label{sec5}
The training datasets can be found at \href{CelebA-dataset}{https://mmlab.ie.cuhk.edu.hk/projects/CelebA.html} and \href{CelebAHQ-dataset}{https://www.kaggle.com/datasets/lamsimon/celebahq}. 

The code relative to this work is available on Github in the following repository:
 \href{https://github.com/asperti/We_love_latent_space}{https://github.com/asperti/We_love_latent_space}. We also provide pretrained weights that can be downloaded using suitable facilities. 
\bigskip

\subsection*{Acknowledgements} We would like to thank Fabio Merizzi for many interesting discussions on the subject of this article.\bigskip

\subsection*{Conflict of interest} On behalf of all authors, the corresponding author states that there is no conflict of interest.

\bibliography{bibliography}

\appendix
\section{Models}
\label{appendix:models}
In this section we briefly discuss the architecture of the generative models used for our experiments. In addition, we also largely experimented with StyleGAN, whose structure is discussed in Section\ref{sec:gans}. Two of the models are vanilla implementations of
GAN and VAE with very similar structures; this was an intentional choice since we wished
to evaluate the impact of the objective function independently from the network architecture.

\subsection{Vanilla GAN Structure}

The structure of the discriminator is as follows:
\begin{enumerate}
    \item A Convolutional layer going from an input of size $(64, 64, 3)$ with stride $s = 2$, same padding, ReLU activation, kernel size $k = 4$ and $128$ channels, followed by a Leaky ReLU layer with $\alpha = 0.2$ for regularization;
    \item A Convolutional layer as in 1. but with $256$ channels, followed by another leaky ReLU;
    \item A Convolutional layer as in 1. but with $512$ channels, followed by another leaky ReLU;
    \item A Dropout layer with $\alpha = 0.2$ for GAN regularization;
    \item A Dense layer outputting a single value, which is the confidence the discriminator has that its input image is real.
\end{enumerate}

The structure of the generator is instead the following:
\begin{enumerate}
    \item A Dense layer going from $L$ to size $(8, 8, 16)$;
    \item A Transposed Convolutional layer with stride $s = 2$, same padding, ReLU activation, kernel size $k = 4$ and $128$ channels, followed by a Leaky ReLU layer with $\alpha = 0.2$ for regularization;
    \item A Transposed Convolutional layer as in 1. but with $256$ channels, followed by another leaky ReLU;
    \item A Transposed Convolutional layer as in 1. but with $512$ channels, followed by another leaky ReLU;
    \item A Convolutional layer with $3$ channels, kernel size $k = 5$, sigmoid activation and same padding, thus producing a $(64, 64, 3)$ output.
\end{enumerate}

We also implemented a re-coder, for GAN inversion, with an essentially symmetric structure.

\subsection{Vanilla VAE Structure}
Our VAEs use a balancing $\gamma$ factor for its two loss components, which are the KL divergence and the reconstruction error, as suggested in \cite{vae_loss_balancing} in order to improve variability and reduce blurriness.

The structure of the encoder is as follows:
\begin{enumerate}
    \item A Convolutional layer going from an input of size $(64, 64, 3)$ with stride $s = 2$, same padding, ReLU activation, kernel size $k = 4$ and $128$ channels, followed by a Leaky ReLU layer with $\alpha = 0.2$ for regularization;
    \item A Convolutional layer as in 1. but with $256$ channels, followed by another leaky ReLU;
    \item A Convolutional layer as in 1. but with $512$ channels, followed by another leaky ReLU;
    \item A Dropout layer with $\alpha = 0.2$ for GAN regularization;
    \item Two separate Dense layers corresponding to the mean and variance vectors of the latent space gaussian of a sample with sizes equal to $L$, respectively, plus a third non-trainable layer which performs the sampling from that same gaussian to obtain a latent vector of size $L$.
\end{enumerate}

The structure of the decoder is the same as the structure of the GAN generator.

\subsection{SVAE Structure}
Split-VAE (SVAE) \cite{SVAE} is a simple architectural variation of a traditional vae where the output $\hat{x}$ is computed as a weighted sum 
\[\hat{x}=\sigma \odot \hat{x_1} + (1-\sigma) \odot \hat{x_2}\] 
of two generated images $\hat{x_1},\hat{x_2}$, and
a {\em learned} compositional map $\sigma$. The splitting structure facilitate the synthesis of uncorrelated latent features, usually permitting to work with latent spaces of higher dimension.

For the implementation of the encoder and the decoder we adopted a 
ResNet-like architecture derived from \cite{TwoStage} that we already used in previous works \cite{balancing,VAEGreen}. 
The basic component, used both for encoding and decoding is
a Scale-block, described in Figure~\ref{fig:Scale-Block}.

\begin{figure}[ht]
\begin{center}
\includegraphics[width=.33\textwidth]{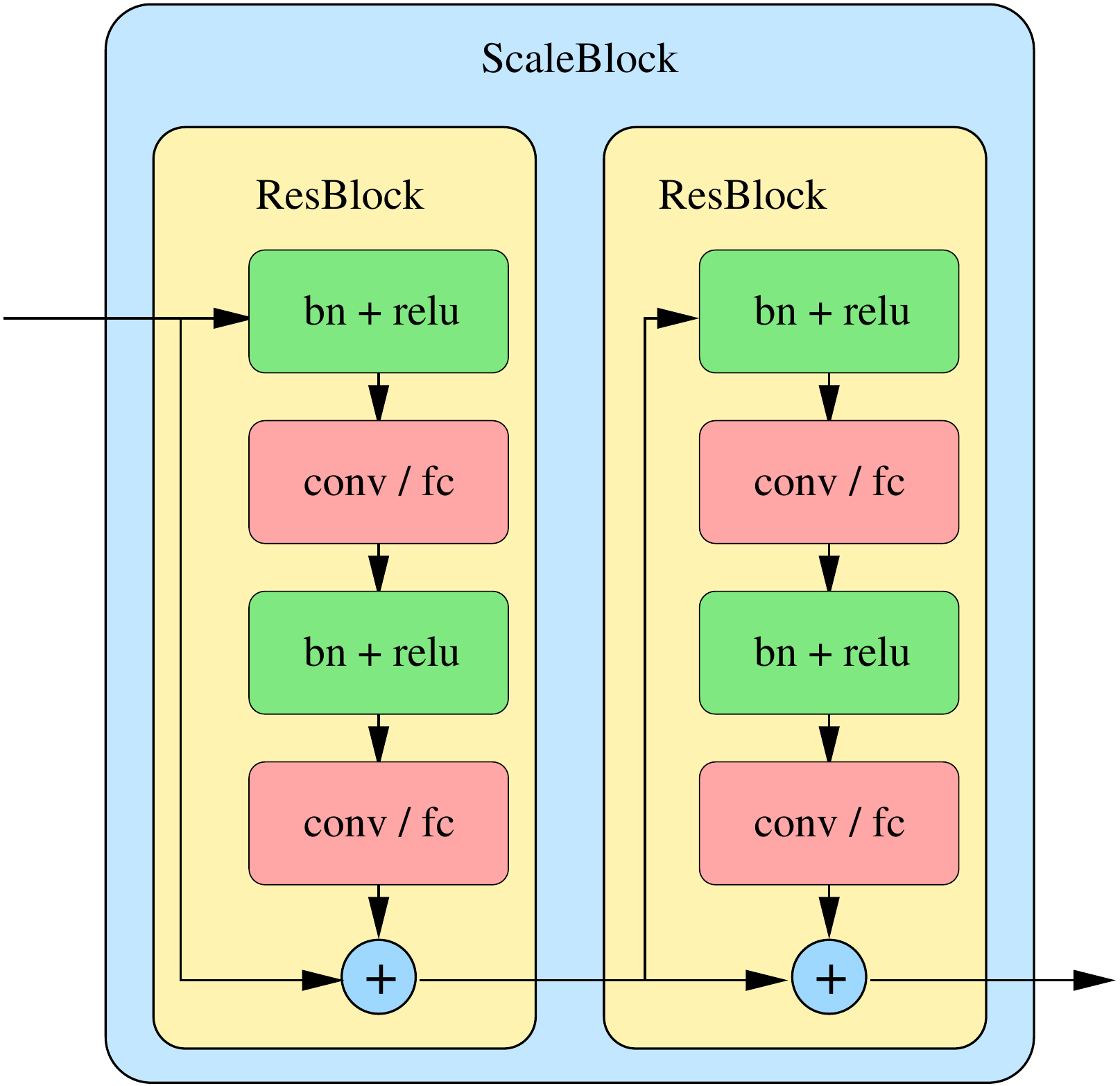}
\end{center}
\caption{Scale Block: a Scale Block is a sequence of Residual Blocks intertwined with residual connections. A Residual Block alternates BatchNormalization layers, non-linear units and convolutions.
\label{fig:ScaleBlock}}
\end{figure}

\begin{figure}[ht]
\begin{center}
\begin{tabular}{cc}
\includegraphics[width=.16\textwidth]{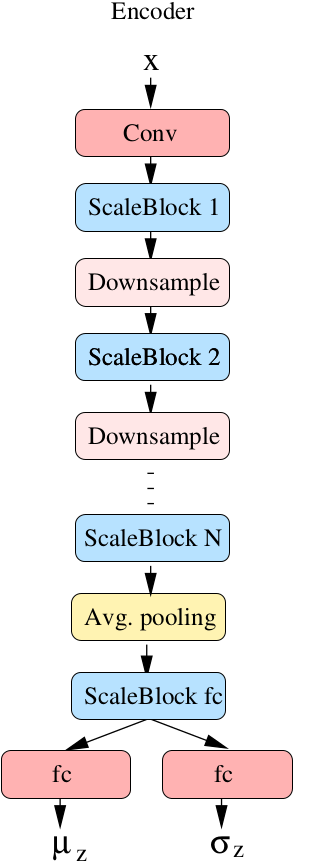} &
\includegraphics[width=.15\textwidth]{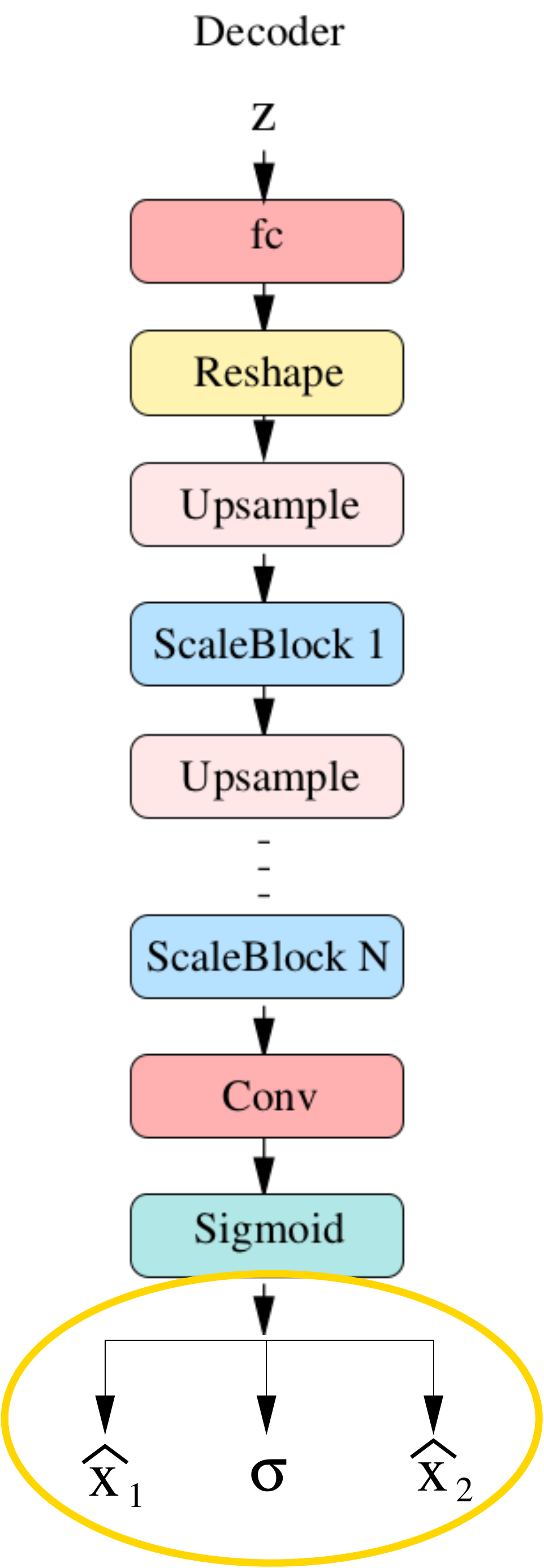}\\
Encoder & Decoder
\end{tabular}
\caption{Encoder: the input is progressively downsampled via convolutions, preceded by Scale Blocks. At the final scale, a global average pooling layer extract features that are further processed via dense layers to compute mean and variance for latent variables. Decoder: the decoder is essentially symmetric. A SVAE only differs in the final layer (circled in the picture): instead of directly producing $\hat{x}$, it produces two images $\hat{x_1}$ and $\hat{x_2}$ and a compositional map $\sigma$, defining $\hat{x}=\sigma\odot\hat{x_1}+(1-\sigma)\odot\hat{x_2}$.}
\label{fig:Resnet}
\end{center}
\end{figure}

Encoder and decoder are essentially alternations of ScaleBlocks and downsampling/upsampling layers, as described in Figure~\ref{fig:Resnet}.

\section{Labels for the support set}
\label{appendix:labels}
In this section we give the list of labels for elements in the support set that we
used for our experiments. The precise set is not very relevant; other choices 
driven by the methodology described in section give similar results. 

Due to memory limitations, we have been forced to restict the investigation 
to the first
70000 images in the CelebA dataset. This is the full list (150 elements):

\noindent
[30, 58, 298, 702, 842, 873, 1779, 1809, 1844, 2590, 2719, 3888, 4114, 4223, 4550, 5659, 5718, 6058, 6108, 6128, 6175, 6244, 6705, 6815, 7499, 7679, 8225, 9457, 11254, 11282, 12367, 13077, 13371, 13993, 14193, 15390, 15711, 15817, 16505, 17186, 17458, 18250, 18283, 18582, 19080, 19175, 19612, 22505, 22633, 23173, 23199, 23308, 23511, 24231, 26431, 27169, 28270, 28401, 28433, 29453, 30248, 30269, 30619, 31741, 31795, 31836, 31978, 32272, 32770, 32828, 33332, 33613, 33669, 34024, 35804, 35823, 35882, 35944, 36483, 36926, 37374, 37534, 37538, 37572, 37682, 38194, 38483, 38677, 39232, 39267, 39424, 39901, 40405, 41464, 42969, 43035, 43199, 44054, 44252, 44589, 44798, 45930, 46259, 46693, 48128, 48786, 48839, 49498, 50345, 52454, 52516, 52673, 52753, 52834, 53071, 53308, 54937, 56128, 56492, 56693, 57844, 57927, 57942, 58020, 58089, 58162, 58389, 58947, 60359, 61004, 61180, 61374, 61495, 61530, 61794, 61878, 63535, 63891, 64328, 64342, 64663, 65041, 66277, 66321, 66663, 68027, 68753, 69274, 69750, 69936]

\begin{figure}[ht]
\includegraphics[width=\columnwidth]{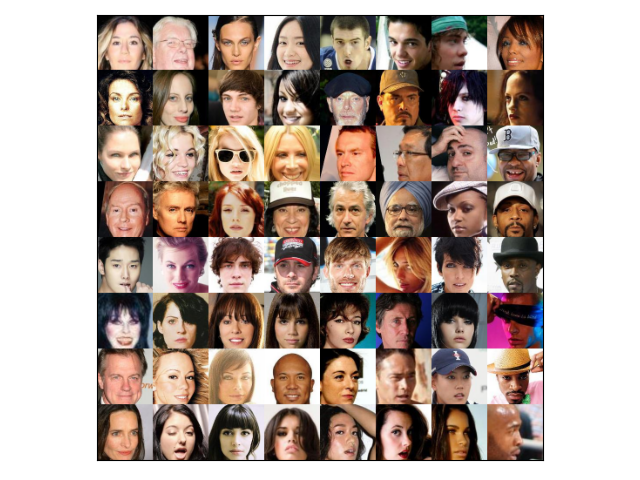}
\caption{Examples of images in the support set, in addition to those in
Figure\ref{fig:support_set1}.
\label{fig:support_set2}}
\end{figure}

As it is clear from Figure~\ref{fig:support_set2} some images in the support set
are a bit pathological: extreme poses, frequent use of accessories like hats and eyeglasses, strange illumination, etc. So the support set also provides a good
test-bench to check (through inversion) the robustness and diversification of the
generative model. 

If required, a more ``comformist'' Support Set can be easily derived by reducing the threshold constraint as in Section \ref{ss:sample_selection}.

\end{document}